\documentclass[10pt,twocolumn,letterpaper]{article}

 \usepackage[pagenumbers]{cvpr} 

\definecolor{cvprblue}{rgb}{0.21,0.49,0.74}
\usepackage[pagebackref,breaklinks,colorlinks,allcolors=cvprblue]{hyperref}

\def\paperID{AIMS-5} 
\def\confName{CVPR}
\def\confYear{2026}

\title{The Deployment Gap in AI Media Detection: Platform-Aware and Visually Constrained Adversarial Evaluation}

\author{
Aishwarya Budhkar\\
Indiana University\\
{\tt\small abudhkar@iu.edu}
\and
Trishita Dhara\\
Upper Hand\\
{\tt\small trishitadhara123@gmail.com}
\and
Siddhesh Sheth\\
Ace Rent a Car\\
{\tt\small shethsiddhesh268@gmail.com}
}

\begin{document}
\maketitle
\begin{abstract}
Recent AI media detectors report near-perfect performance under clean laboratory evaluation, yet their robustness under realistic deployment conditions remains underexplored. In practice, AI-generated images are resized, compressed, re-encoded, and visually modified before being shared on online platforms. We argue that this creates a deployment gap between laboratory robustness and real-world reliability.

In this work, we introduce a platform-aware adversarial evaluation framework for AI media detection that explicitly models deployment transforms (e.g., resizing, compression, screenshot-style distortions) and constrains perturbations to visually plausible meme-style bands rather than full-image noise. Under this threat model, detectors achieving AUC $\approx$ 0{.}99 in clean settings experience substantial degradation. Per-image platform-aware attacks reduce AUC to significantly lower levels and achieve high fake-to-real misclassification rates, despite strict visual constraints. We further demonstrate that universal perturbations exist even under localized band constraints, revealing shared vulnerability directions across inputs. Beyond accuracy degradation, we observe pronounced calibration collapse under attack, where detectors become confidently incorrect.

Our findings highlight that robustness measured under clean conditions substantially overestimates deployment robustness. We advocate for platform-aware evaluation as a necessary component of future AI media security benchmarks and release our evaluation framework to facilitate standardized robustness assessment.
\end{abstract}    
\section{Introduction}

Recent advances in generative models have enabled the creation of highly realistic synthetic media, raising urgent concerns about misinformation, impersonation, and media manipulation. In response, a growing body of work has proposed AI media detectors that distinguish real from synthetic content, often reporting near-perfect performance under clean laboratory evaluation \cite{wang2020cnn, ojha2023towards}. However, such evaluations typically assume access to pristine images and do not model the transformations that occur during real-world content sharing.

In practice, AI-generated images are rarely consumed in their original form. They are resized, compressed, re-encoded, screenshotted, and visually modified before being posted on social platforms. These transformations alter statistical cues that forensic detectors rely on. Prior work has demonstrated that distribution shifts and corruptions can significantly degrade uncertainty calibration and predictive reliability \cite{ovadia2019can}, while adversarial examples further expose brittleness in deep neural networks \cite{goodfellow2015explaining, madry2018towards}. Yet, the interaction between platform-induced transformations and adversarial robustness in AI media detection remains underexplored.

We argue that this mismatch constitutes a \emph{deployment gap}: a discrepancy between laboratory robustness and deployment robustness. Existing adversarial studies typically optimize perturbations in the pixel space under clean evaluation assumptions. Meanwhile, media forensics benchmarks often ignore adversarial threat models altogether. As a result, reported performance may substantially overestimate robustness under realistic platform conditions.

In this work, we introduce a \textbf{platform-aware adversarial evaluation framework} for AI media detection. Our framework incorporates differentiable approximations of common deployment transforms (e.g., resizing, compression, screenshot-style distortions) into the attack loop, enabling end-to-end optimization under realistic sharing conditions. Furthermore, we constrain perturbations to visually plausible meme-style bands rather than full-image noise, reflecting common user-generated modifications. 

Under this threat model, detectors achieving AUC $\approx 0.99$ under clean conditions experience substantial degradation. Platform-aware, band-constrained per-image attacks induce high fake-to-real misclassification rates and significantly reduce AUC. We further demonstrate that universal perturbations exist even under localized visual constraints, revealing shared vulnerability directions across inputs. Beyond accuracy degradation, we observe pronounced calibration collapse under attack, with detectors becoming confidently incorrect under deployment-like conditions.

Our findings suggest that clean laboratory evaluation substantially overestimates real-world robustness. We advocate for platform-aware robustness assessment as a necessary component of future AI media security benchmarks and release our evaluation framework to facilitate standardized deployment-aware testing. Code available at : https://github.com/trishitadhara/Platform-Aware-and-Visually-Constrained-Adversarial-Evaluation

\section{Related Work}

\subsection{AI-Generated Media Detection}

The rapid progress of generative models has led to extensive research on detecting synthetic images. Early studies showed that CNN-generated images contain detectable artifacts in spatial statistics and frequency spectra \cite{wang2020cnn, durall2020watch}. More recent work has focused on generalization across generators, aiming to detect images produced by unseen models \cite{ojha2023towards, coccomini2022combining}. Patch-level consistency and universal detection approaches have also been explored \cite{liu2022universal}. 

While these methods report strong performance under clean evaluation, they are typically assessed using pristine images and standard test splits. Robustness under realistic deployment conditions, including platform-induced transformations, remains less systematically examined.

\subsection{Adversarial Robustness}

Adversarial examples reveal the vulnerability of deep neural networks to small perturbations \cite{goodfellow2015explaining}. Robust optimization frameworks \cite{madry2018towards} and universal adversarial perturbations \cite{moosavi2017universal} have demonstrated structural weaknesses in learned decision boundaries. Athalye et al.~\cite{athalye2018obfuscated} introduced expectation-over-transformation (EOT), enabling attacks to remain effective under stochastic transformations.

Although adversarial robustness has been widely studied in classification tasks, its intersection with AI media detection remains underdeveloped. Existing adversarial evaluations of forensic systems often rely on unconstrained perturbations or do not explicitly incorporate realistic platform transformations.

\subsection{Robustness Under Distribution Shift and Corruptions}

Beyond adversarial perturbations, neural networks are sensitive to natural distribution shifts and corruptions \cite{hendrycks2019benchmarking}. Predictive uncertainty has been shown to degrade significantly under shift \cite{ovadia2019can}, motivating robustness benchmarks such as ImageNet-C and WILDS \cite{taori2020measuring, koh2021wilds}. 

However, most distribution shift studies focus on natural corruptions rather than adversarial manipulations optimized under deployment pipelines. The combination of platform transformations and constrained adversarial optimization remains insufficiently studied in media security contexts.

\subsection{Positioning and Novelty}

Existing work in AI media detection has primarily focused on improving detection accuracy under clean evaluation settings \cite{wang2020cnn, ojha2023towards}. Meanwhile, adversarial robustness research has demonstrated the vulnerability of deep networks to carefully optimized perturbations \cite{madry2018towards, moosavi2017universal}. However, these two research directions have largely evolved in parallel.

Our work bridges this gap by explicitly modeling deployment conditions within adversarial evaluation for AI media detection. Unlike prior adversarial studies that assume direct access to pristine inputs, we incorporate differentiable approximations of common platform transformations into the attack loop. Furthermore, we constrain perturbations to visually plausible meme-style bands, reflecting realistic user-generated modifications rather than unconstrained full-image noise.

Rather than proposing a new detector or defense, we provide a deployment-aware robustness evaluation framework. Our findings demonstrate that clean laboratory robustness substantially overestimates performance under realistic sharing conditions. We also show that universal perturbations exist even under localized visual constraints, revealing shared vulnerability directions in detection models.

By formalizing and empirically analyzing this deployment gap, our work contributes a practical robustness assessment framework for AI media security research.

\section{Threat Model and Methodology}

\subsection{Deployment-Aware Threat Model}

We consider the problem of AI-generated image detection under realistic deployment conditions. Let $f_\theta(x)$ denote a detector that outputs the probability that an image $x$ is synthetic. Standard adversarial evaluation optimizes perturbations directly on $x$ under pixel-level constraints. However, in real-world settings, images are rarely evaluated in pristine form.

We model a deployment pipeline $\mathcal{T}$ consisting of transformations that approximate common platform processing operations, including resizing, compression, re-encoding, and screenshot-style distortions. Our attacker optimizes perturbations through this pipeline, seeking to minimize detector confidence under transformed inputs:
\[
\min_{\delta} \; \mathbb{E}_{t \sim \mathcal{T}} \left[ \ell(f_\theta(t(x + \delta)), y_{\text{target}}) \right],
\]
subject to perceptual and structural constraints.

\subsection{Visually Constrained Perturbations}

Unlike conventional full-image perturbations \cite{goodfellow2015explaining, madry2018towards}, we restrict $\delta$ to localized, visually plausible regions. Specifically, perturbations are confined to meme-style bands (e.g., top or bottom strips of the image), reflecting common user-generated modifications on social platforms. This constraint ensures that adversarial modifications remain consistent with real-world sharing behavior and avoids unrealistic full-frame noise patterns.

\subsection{Per-Image and Universal Attacks}

We study two attack regimes:

\textbf{Per-image attacks}, where a unique perturbation is optimized for each input, and

\textbf{Universal perturbations}, where a single perturbation is optimized across multiple inputs \cite{moosavi2017universal}. Universal perturbations are of particular interest, as they reveal shared vulnerability directions in the detector's decision boundary.

Both attack types are optimized using projected gradient descent with expectation over transformations (EOT) \cite{athalye2018obfuscated} to account for stochastic deployment transforms.

\begin{figure}[t]
\centering
\includegraphics[width=\columnwidth]{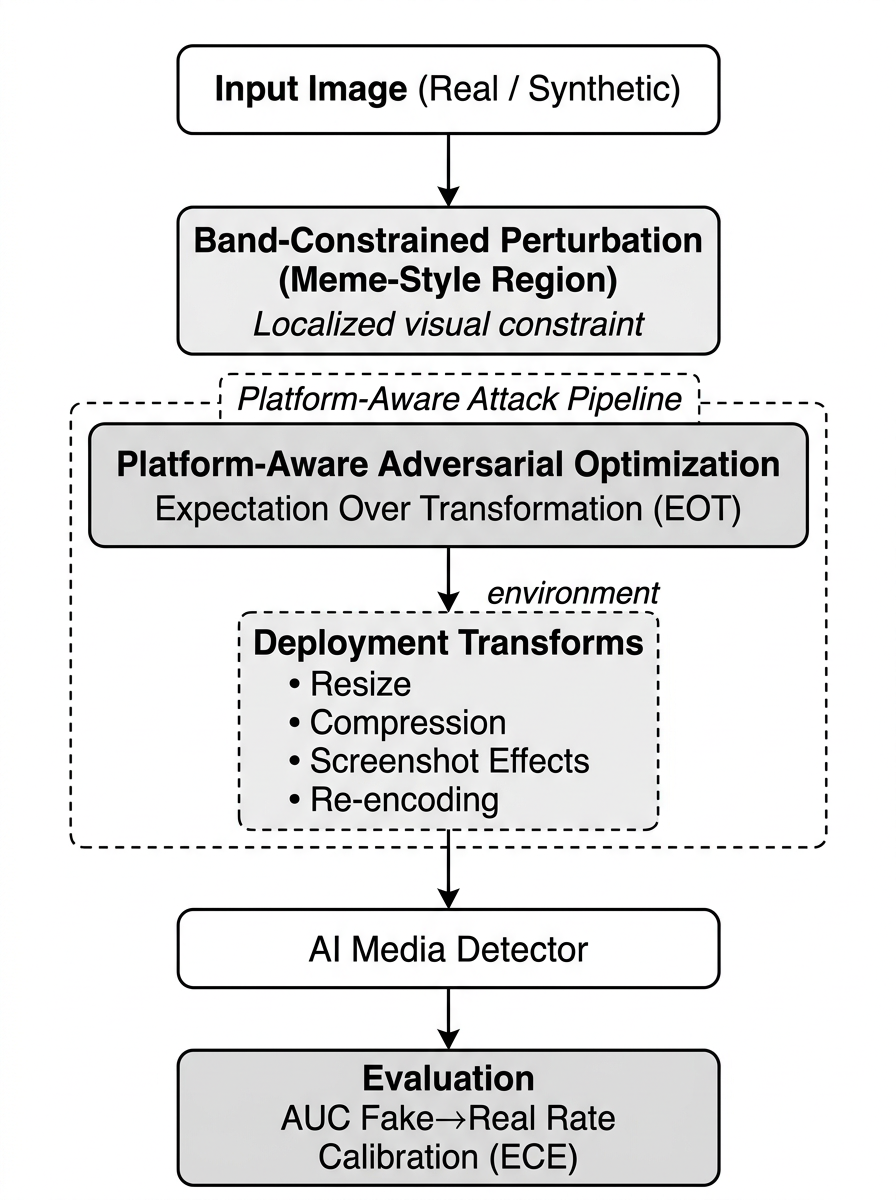}
\caption{
Overview of the deployment-aware adversarial evaluation framework. 
Perturbations are constrained to visually plausible meme-style bands and optimized through platform transformations using expectation-over-transformation (EOT). 
Detector robustness is evaluated using AUC, misclassification rate, and calibration metrics.
}
\label{fig:methodology}
\end{figure}

An overview of the proposed deployment-aware evaluation framework is shown in Figure~\ref{fig:methodology}.

\section{Experiments}

\subsection{Dataset}

We construct a balanced real-vs-synthetic dataset. Real images are drawn from the STL-10 dataset \cite{coates2011analysis}, a natural image benchmark commonly used in representation learning. Synthetic images are generated using Stable Diffusion \cite{rombach2022high} across 10 diverse prompts designed to mimic realistic photographic scenarios (e.g., urban street scenes, wildlife photography, sports events, portraits, protests, and travel imagery). Each prompt is assigned a unique identifier (\texttt{prompt\_id}) to enable per-prompt robustness analysis. For each prompt, 100 synthetic images are generated using different random seeds to ensure diversity. Images are resized to $224 \times 224$ resolution. The dataset is split into training and test partitions, with synthetic images evenly distributed across prompts to enable per-prompt analysis.

\subsection{Detectors}

We evaluate two detector architectures.

\textbf{CLIP-based detector.}
We adopt a CLIP ViT-based backbone \cite{radford2021learning} with a linear classification head trained to distinguish real from synthetic images. The backbone is frozen and only the classification head is optimized.

\textbf{ResNet-18 detector.}
We fine-tune a ResNet-18 classifier \cite{he2016deep} on the same dataset.

Under clean evaluation, both detectors achieve near-perfect performance (AUC $\approx 0.99$), consistent with prior reports of strong clean detection accuracy.

\subsection{Attack Settings}

We evaluate both per-image and universal adversarial perturbations under a deployment-aware threat model.

\paragraph{Per-image attacks.}
For each synthetic image, we optimize a perturbation constrained to a meme-style band region (top or bottom of the image). Optimization is performed using projected gradient descent (PGD) \cite{madry2018towards} with expectation over transformation (EOT) \cite{athalye2018obfuscated} to account for stochastic deployment transforms. The perturbation is bounded in $\ell_\infty$ norm and spatially restricted to the designated band region.

\paragraph{Universal perturbations.}
We additionally optimize a single perturbation shared across multiple synthetic images, following the formulation of universal adversarial perturbations \cite{moosavi2017universal}. Universal perturbations are trained using the same deployment-aware EOT framework.

\paragraph{Deployment transforms.}
The transformation set includes resizing, JPEG-style compression, and screenshot-like distortions, inspired by corruption robustness benchmarks \cite{hendrycks2019benchmarking}. These operations are incorporated during optimization to simulate realistic platform processing.

\subsection{Evaluation Metrics}

We report:

\begin{itemize}
    \item \textbf{AUC}: Area Under the ROC Curve.
    \item \textbf{Fake-to-Real Misclassification Rate}: proportion of synthetic images misclassified as real.
    \item \textbf{Expected Calibration Error (ECE)} \cite{guo2017calibration}: to measure reliability under attack.
\end{itemize}

\section{Results}

\subsection{Clean vs Deployment-Aware Performance}

\begin{figure}[!ht]
\centering

\begin{subfigure}{\columnwidth}
    \centering
    \includegraphics[width=0.8\linewidth]{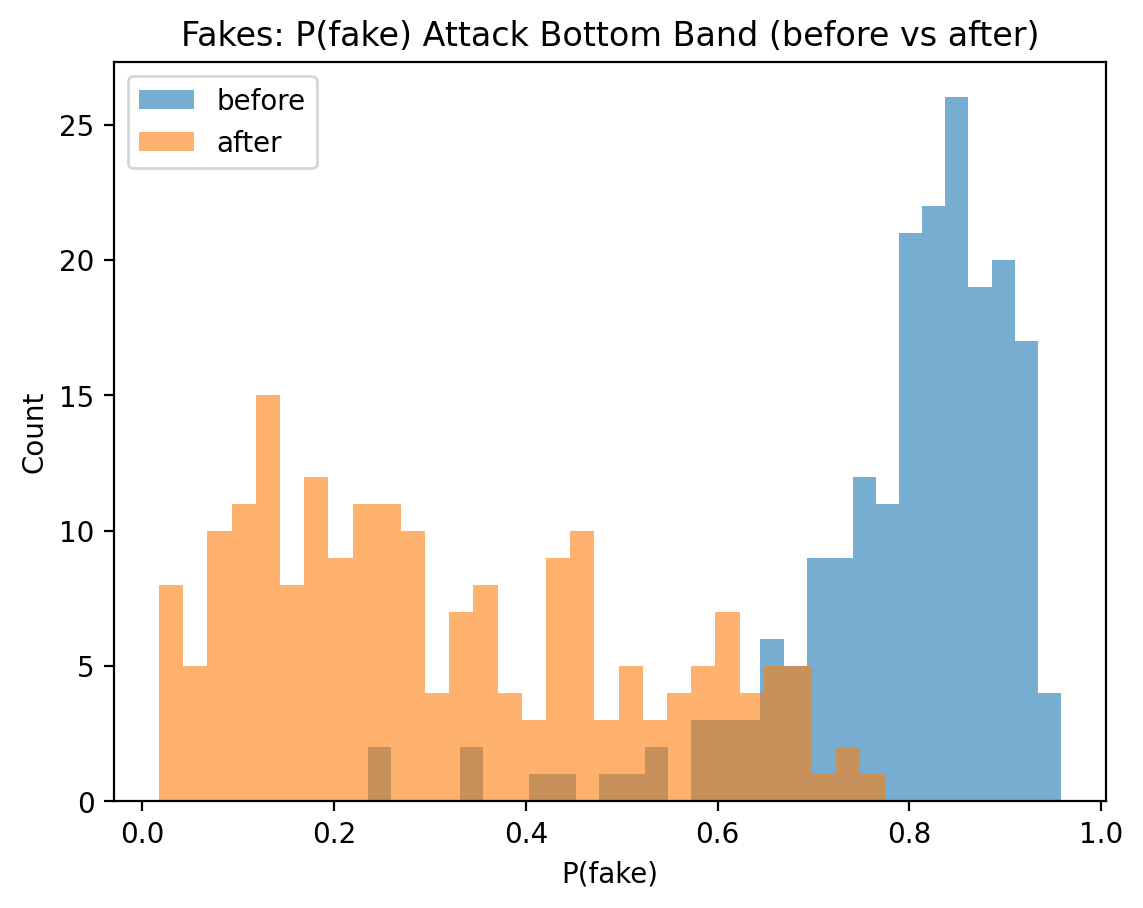}
    \caption{Bottom-band attack}
\end{subfigure}

\vspace{0.5em}

\begin{subfigure}{\columnwidth}
    \centering
    \includegraphics[width=0.8\linewidth]{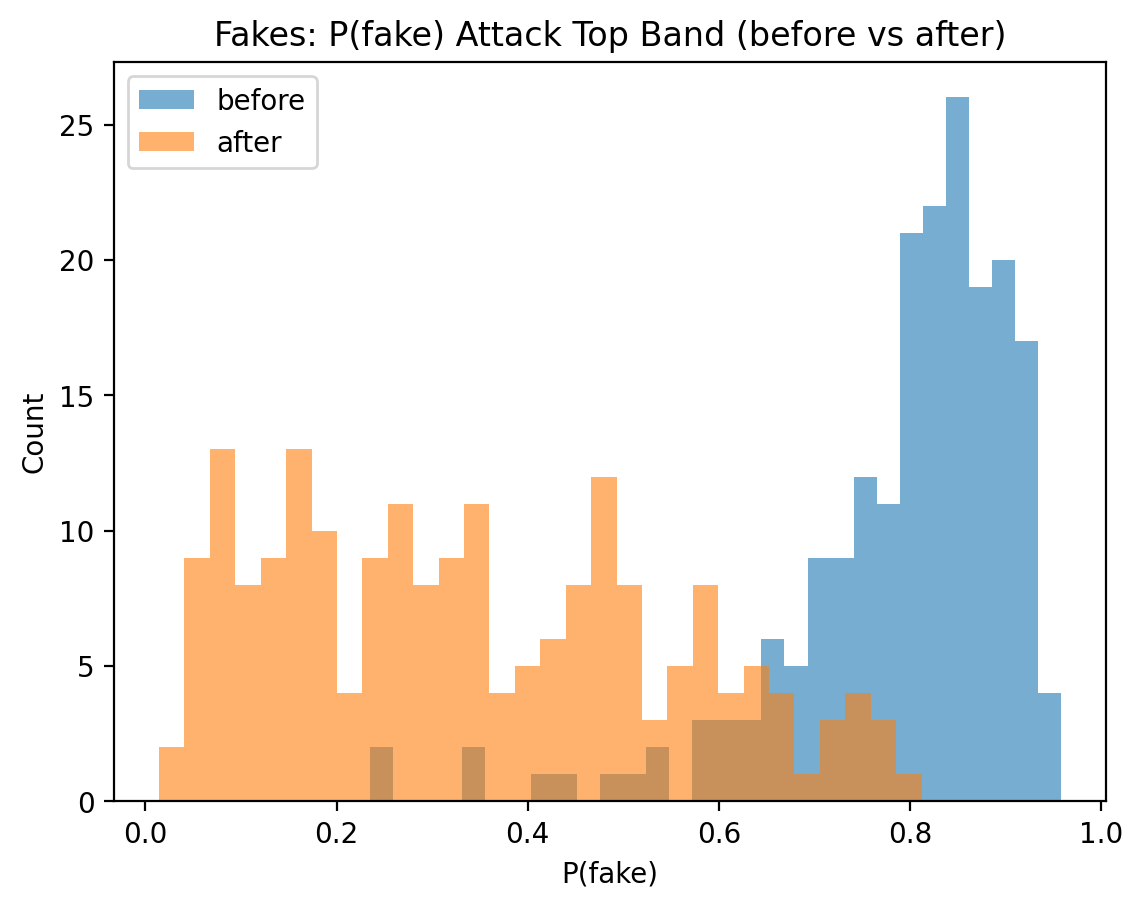}
    \caption{Top-band attack}
\end{subfigure}

\caption{
Confidence distribution for synthetic images under deployment-aware band-constrained attacks.
Both top and bottom perturbations significantly shift detector confidence toward the real class.
}
\label{fig:confidence_shift}
\end{figure}

Figure~\ref{fig:confidence_shift} visualizes confidence distributions for synthetic images under clean and attacked settings. 
Under clean evaluation, synthetic images receive high-confidence fake predictions. 
Under deployment-aware perturbations, the confidence distribution shifts substantially toward the real class, indicating severe robustness degradation.

Table~\ref{tab:overall_results} summarizes detector performance under clean and deployment-aware adversarial conditions. 
Under clean evaluation, the CLIP-based detector achieves AUC $=0.996$ and accuracy $=0.97$. 
However, under platform-aware band-constrained attack (bottom band), AUC drops to $0.705$, and the fake-to-real misclassification rate increases to $0.78$.

\begin{table}[!ht]
\centering
\caption{Overall performance under deployment-aware attack (bottom band).}
\begin{tabular}{lccc}
\toprule
Setting & AUC & Accuracy & Fake$\rightarrow$Real Rate \\
\midrule
Clean & 0.996 & 0.970 & 0.055 \\
Per-Image Attack & 0.705 & 0.885 & 0.780 \\
Universal Attack & 0.923 & 0.825 & 0.345 \\
\bottomrule
\end{tabular}
\label{tab:overall_results}
\end{table}

\subsection{Universal Perturbations}

We further evaluate universal perturbations optimized under the same deployment-aware threat model. 
Despite strict spatial constraints, a single perturbation achieves a fake-to-real misclassification rate of approximately $0.35$, reducing AUC to $0.923$.

\begin{figure}[!ht]
\centering
\includegraphics[width=0.8\columnwidth]{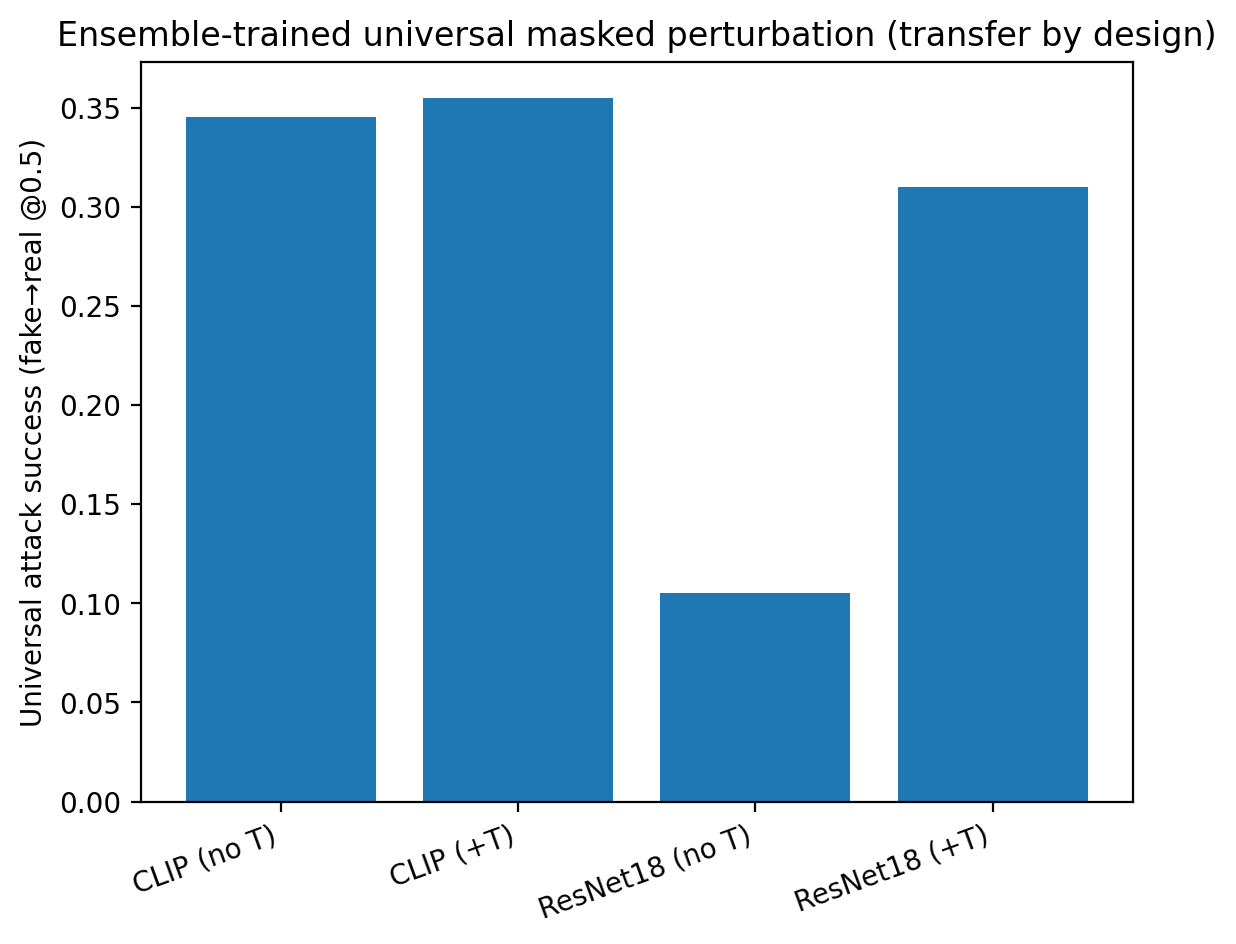}
\caption{
Comparison of per-image and universal platform-aware attacks under band constraints.
}
\label{fig:universal_results}
\end{figure}

Figure~\ref{fig:universal_results} compares per-image and universal attack performance. 
While universal attacks are weaker than per-image optimization, they demonstrate the existence of shared vulnerability directions.

\subsection{Per-Prompt Vulnerability}

\begin{figure}[!ht]
\centering

\begin{subfigure}{\columnwidth}
    \centering
    \includegraphics[width=\linewidth]{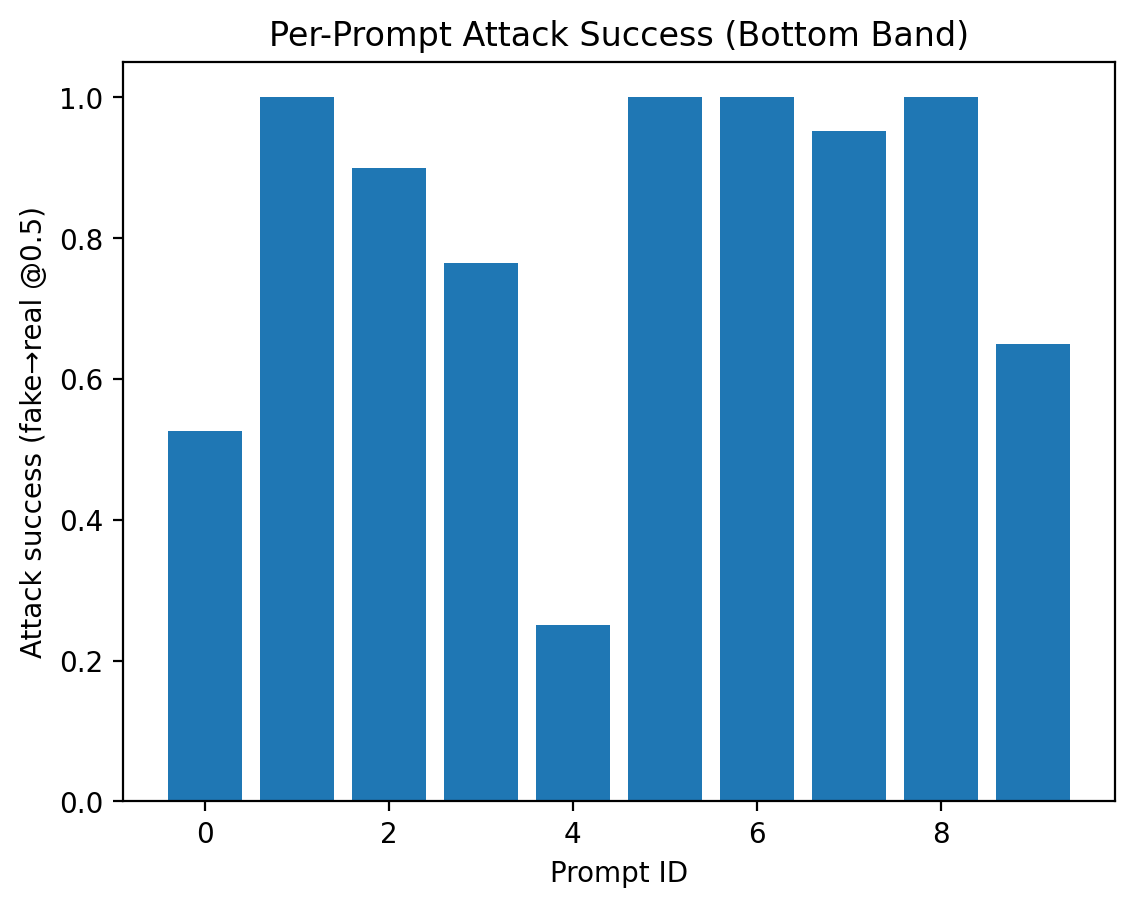}
    \caption{Bottom-band attack}
\end{subfigure}

\vspace{0.4em}

\begin{subfigure}{\columnwidth}
    \centering
    \includegraphics[width=\linewidth]{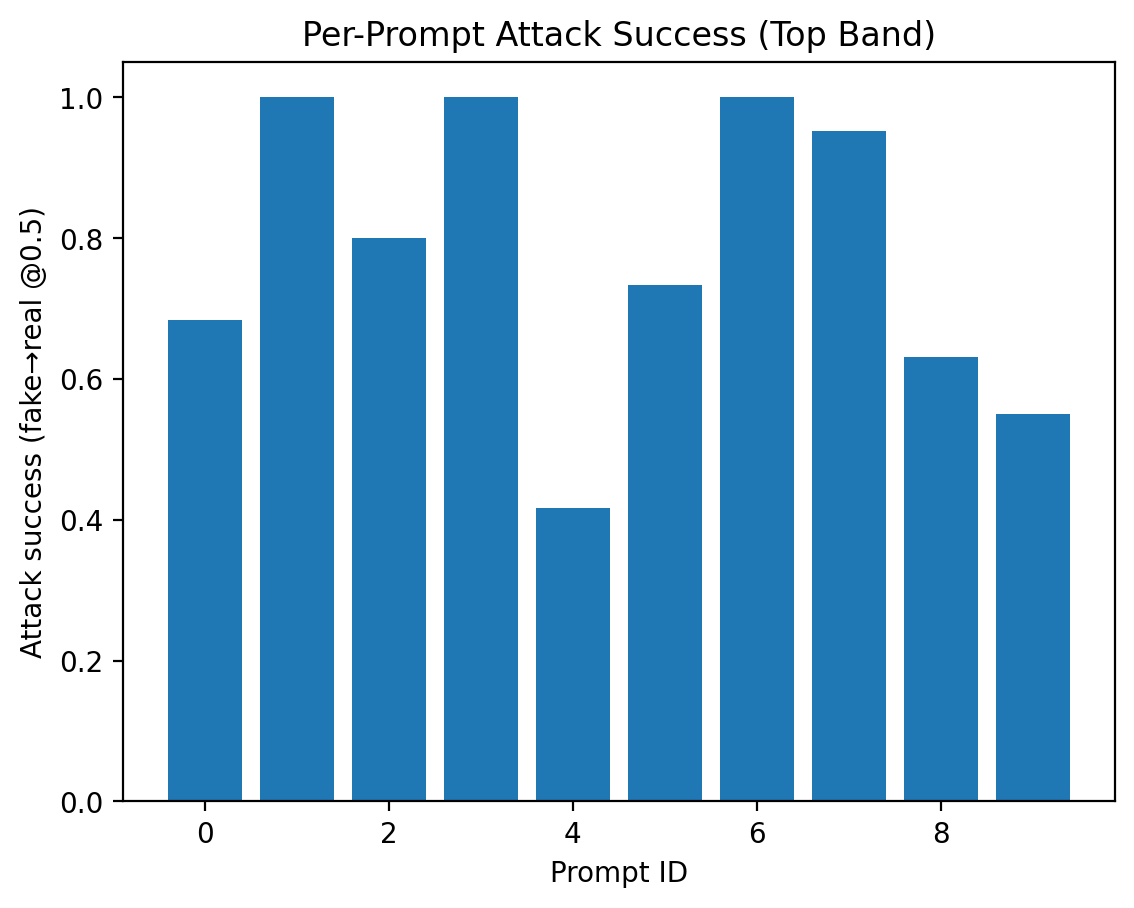}
    \caption{Top-band attack}
\end{subfigure}

\caption{
Fake-to-real misclassification rate per prompt category under deployment-aware band-constrained attacks. 
Vulnerability varies across semantic prompt categories for both top and bottom perturbation regions.
}
\label{fig:per_prompt_results}
\end{figure}

To analyze semantic sensitivity, we evaluate attack success per prompt category. 
Figure~\ref{fig:per_prompt_results} shows fake-to-real misclassification rates grouped by \texttt{prompt\_id}. 
We observe variability across prompts, suggesting that robustness differs across semantic image categories.

\subsection{Calibration Collapse}

Beyond accuracy degradation, we evaluate calibration using Expected Calibration Error (ECE). 
Table~\ref{tab:ece_results} reports ECE under clean and attacked settings. 
Under deployment-aware perturbations, calibration error increases substantially, indicating that the detector becomes confidently incorrect.

\begin{table}[!ht]
\centering
\caption{Expected Calibration Error (ECE) under clean and attack conditions.}
\begin{tabular}{lc}
\toprule
Setting & ECE \\
\midrule
Clean & 0.16743 \\
Attack (Bottom Band) & 0.25064 \\
Attack (Top Band) & 0.23679 \\
\bottomrule
\end{tabular}
\label{tab:ece_results}
\end{table}

\begin{figure}[!ht]
\centering

\begin{subfigure}{\columnwidth}
    \centering
    \includegraphics[width=\linewidth]{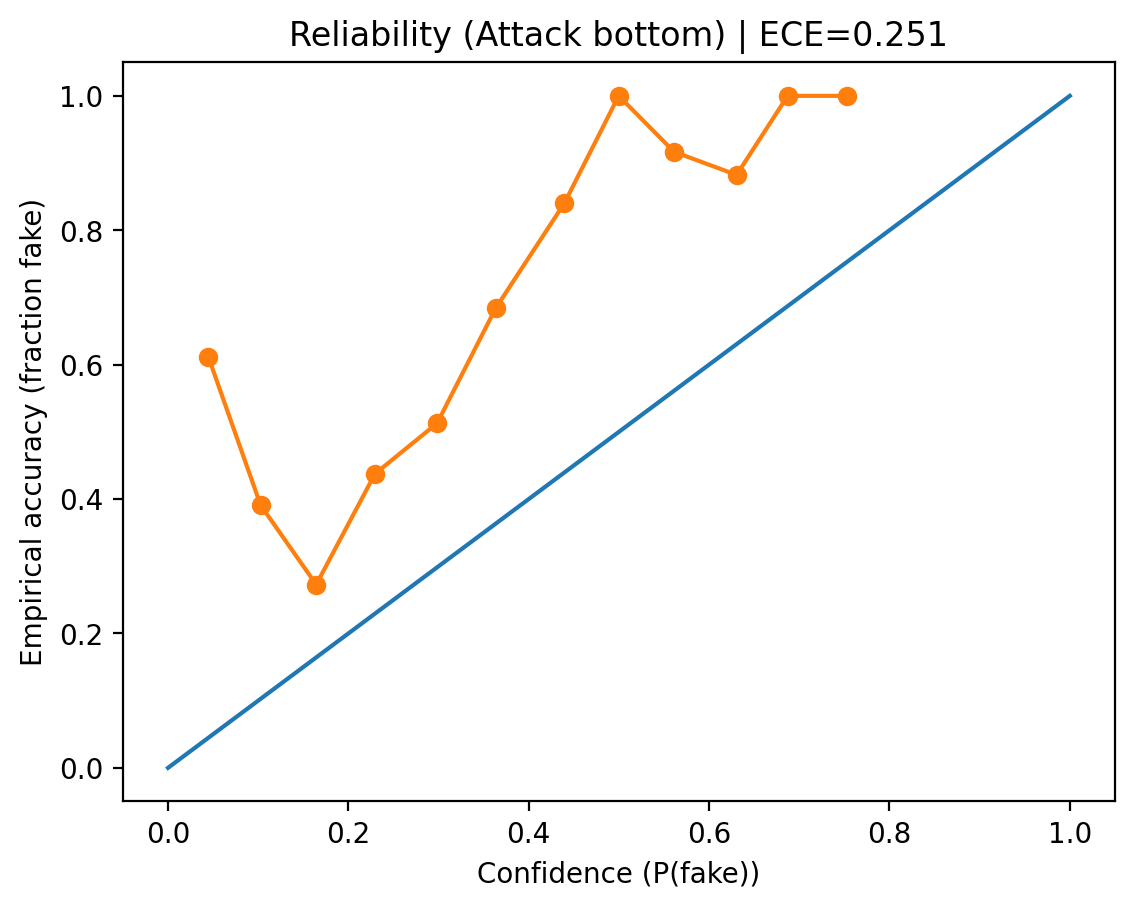}
    \caption{Bottom-band attack}
\end{subfigure}

\vspace{0.5em}

\begin{subfigure}{\columnwidth}
    \centering
    \includegraphics[width=\linewidth]{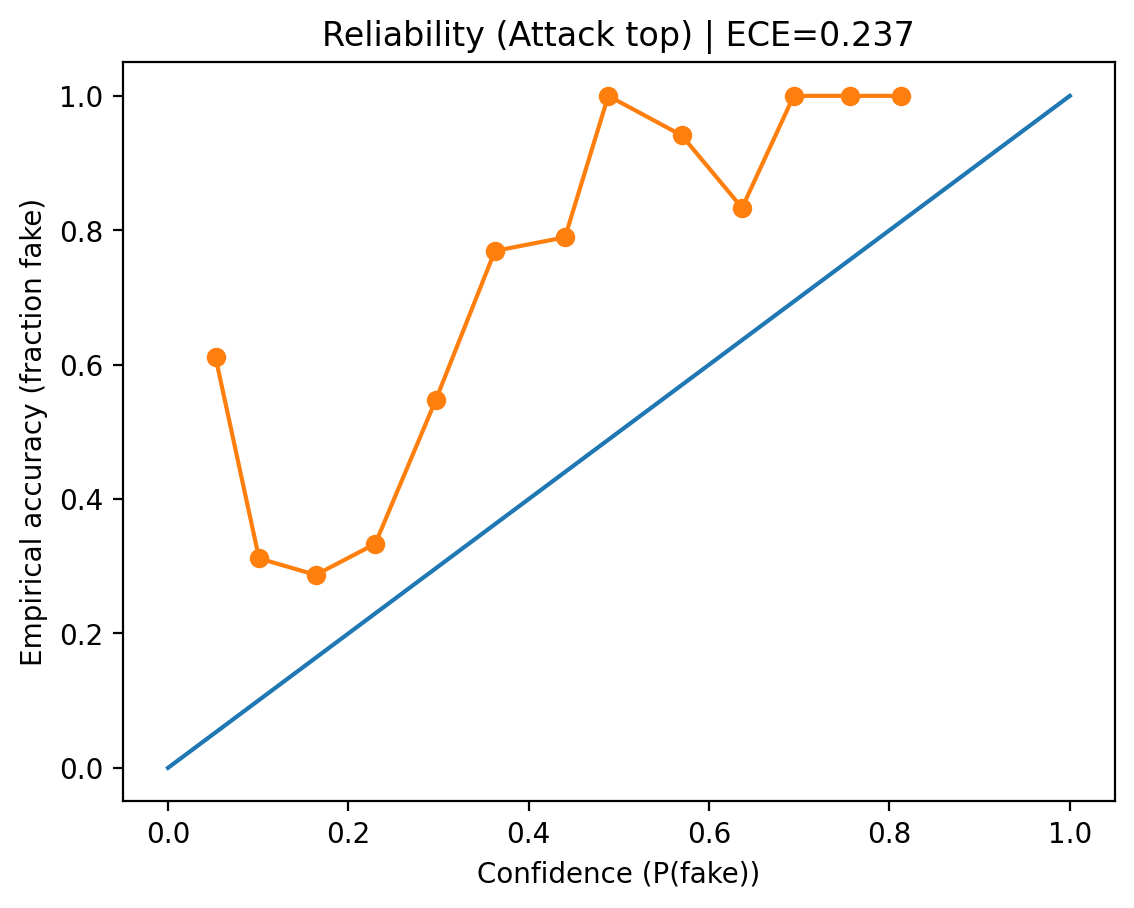}
    \caption{Top-band attack}
\end{subfigure}

\caption{
Reliability diagrams under deployment-aware attacks.
Both perturbation locations induce calibration collapse, with detectors becoming overconfident on misclassified synthetic images.
}
\label{fig:reliability}

\end{figure}

Figure~\ref{fig:reliability} illustrates reliability diagrams under attack, highlighting systematic overconfidence.

\subsection{Statistical Stability via Bootstrap Confidence Intervals}

To assess the statistical stability of our findings, we compute 95\% bootstrap confidence intervals (2{,}000 resamples) for AUC, accuracy, and fake-to-real misclassification rate on the same evaluation slice (CLIP detector, bottom-band attack, deployment-aware evaluation).

Table~\ref{tab:bootstrap_ci} reports the mean and confidence intervals under clean and attacked conditions. 
Under clean evaluation, performance is near-perfect with tight confidence bounds. 
Under deployment-aware attack, performance degrades substantially, with AUC decreasing from $0.997$ to $0.938$ and fake-to-real misclassification increasing from $0.010$ to $0.356$. 

Importantly, the confidence intervals for clean and attack settings do not overlap for key metrics, indicating that the observed degradation is statistically significant and not attributable to sampling variability.

\begin{table}[t]
\centering
\caption{Bootstrap 95\% confidence intervals (2{,}000 resamples) for clean and attacked conditions (CLIP, bottom-band, deployment-aware evaluation).}
\small
\setlength{\tabcolsep}{4pt} 
\begin{tabular}{lcc}
\toprule
Metric & Clean & Attack \\
\midrule
AUC & 
$0.997$ [0.993, 1.000] & 
$0.938$ [0.914, 0.961] \\

Accuracy & 
$0.992$ [0.983, 1.000] & 
$0.820$ [0.783, 0.858] \\

Fake$\rightarrow$Real Rate & 
$0.010$ [0.000, 0.026] & 
$0.356$ [0.289, 0.422] \\
\bottomrule
\end{tabular}
\label{tab:bootstrap_ci}
\end{table}

\subsection{Perturbation Characterization and Perceptual Considerations}

We constrain adversarial perturbations under an $\ell_\infty$ norm bound of $\epsilon = 16/255$ in the normalized $[0,1]$ pixel space. 
This corresponds to a maximum per-pixel intensity deviation of approximately 6.3\% of the dynamic range. 
In addition to magnitude constraints, perturbations are spatially restricted to horizontal meme-style bands covering 22\% of the image height (either top or bottom), ensuring localized rather than full-frame modifications.

We do not explicitly optimize perceptual similarity metrics such as LPIPS or SSIM. 
Instead, perceptual plausibility is enforced through the combination of a small $\ell_\infty$ bound and strict spatial localization, reflecting common user-generated overlays and meme-style edits frequently observed on social platforms.

\begin{figure}[!ht]
\centering
\includegraphics[width=0.7\columnwidth]{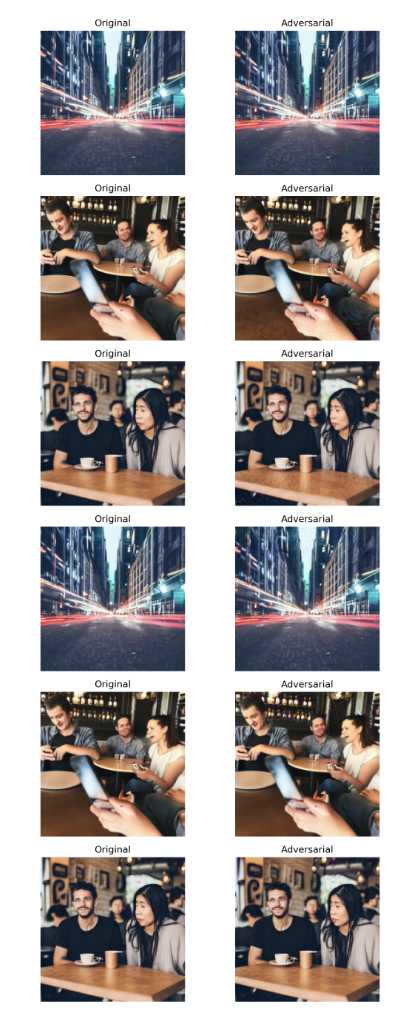}
\caption{
Representative qualitative examples of band-constrained adversarial perturbations.
Each row shows an original synthetic image (left) and its adversarially perturbed counterpart (right).
Perturbations are localized to narrow horizontal bands and bounded by $\epsilon = 16/255$.
}
\label{fig:qualitative_examples}
\end{figure}

Figure~\ref{fig:qualitative_examples} provides representative qualitative examples of perturbed synthetic images. 
Perturbations are confined to narrow horizontal regions and remain visually subtle at standard viewing scale.

We do not conduct a formal human perceptual study. 
While perturbations are designed to be localized and limited in magnitude, future work may incorporate user studies or perceptual similarity metrics to more rigorously evaluate visual realism.
\section{Discussion}

\subsection{The Deployment Gap}

Our results reveal a clear deployment gap in AI media detection. 
While detectors achieve near-perfect performance under clean evaluation (AUC $\approx 0.99$), performance degrades substantially when adversarial perturbations are optimized through realistic platform transformations. 

Importantly, perturbations are spatially restricted to narrow meme-style bands rather than full-image noise. 
Despite these constraints, per-image attacks induce high fake-to-real misclassification rates and significant AUC reduction. 
This suggests that clean laboratory evaluation substantially overestimates robustness under realistic content-sharing conditions.

\subsection{Effect of Platform Transformations}

A key finding is that incorporating deployment transforms directly into the attack loop enables perturbations to remain effective after resizing and compression. 
This highlights that robustness evaluation must account for the full deployment pipeline rather than assuming direct access to pristine inputs. 
Evaluating detectors solely under clean conditions risks providing a misleading sense of security.

\subsection{Universal Vulnerabilities Under Spatial Constraints}

Although universal perturbations are weaker than per-image attacks, their non-trivial success under strict spatial constraints indicates the presence of shared vulnerability directions in the detector’s decision boundary. 
Even localized perturbations can systematically shift model confidence, suggesting that robustness limitations are not purely instance-specific.

\subsection{Calibration and Reliability}

Beyond accuracy degradation, we observe pronounced calibration collapse under deployment-aware attack. 
Detectors become confidently incorrect on perturbed synthetic images, increasing Expected Calibration Error. 
From a media security perspective, this is particularly concerning: overconfident misclassification may mislead automated moderation systems or human reviewers relying on confidence scores.

These findings align with broader evidence that predictive uncertainty degrades under distribution shift, but demonstrate that deployment-aware adversarial conditions exacerbate this effect.

\subsection{Implications for Media Security Evaluation}

Our study suggests that evaluating AI media detectors without modeling realistic sharing conditions may provide an incomplete picture of robustness. 
We advocate for deployment-aware adversarial evaluation as a complementary benchmark dimension alongside clean accuracy and cross-generator generalization.

Rather than proposing new defenses, our work emphasizes the importance of realistic robustness assessment. 
Future detector development should consider deployment-aware adversarial resilience as a first-class objective.

\section{Conclusion}

We introduced a deployment-aware adversarial evaluation framework for AI media detection. 
While modern detectors achieve near-perfect performance under clean laboratory conditions, our results demonstrate substantial robustness degradation when perturbations are optimized through realistic platform transformations and constrained to visually plausible meme-style regions.

We show that deployment-aware per-image attacks significantly reduce AUC and induce high fake-to-real misclassification rates. 
Universal perturbations remain effective under strict spatial constraints, indicating shared vulnerability directions in detector decision boundaries. 
Beyond accuracy degradation, we observe calibration collapse under attack, with detectors becoming confidently incorrect.

Our findings highlight a deployment gap between laboratory evaluation and real-world robustness. 
We argue that platform-aware adversarial testing should become a standard component of AI media security benchmarks to better reflect realistic operating conditions.

\section{Limitations}

Our study has several limitations.

First, synthetic images are generated using a single diffusion-based model. 
Although prompts are diverse, robustness behavior may vary across generative architectures. 
Future work should evaluate detectors across multiple generator families.

Second, our deployment pipeline approximates common platform transformations using differentiable operators. 
While these transformations capture resizing and compression effects, real-world platforms may introduce additional processing steps not modeled here.

Third, we focus on visually constrained band perturbations. 
Other forms of realistic modifications, such as overlays, cropping, or watermark-like artifacts, may exhibit different robustness characteristics.

Finally, our analysis centers on detector robustness rather than defense mechanisms. 
We do not propose mitigation strategies, leaving robustness improvement as an open research direction.

\section{Future Work}

Several directions emerge from our findings.

First, expanding deployment-aware evaluation to multiple generative models and larger-scale datasets would help characterize the generality of the observed deployment gap. 

Second, integrating platform-aware adversarial training could provide insights into whether deployment robustness can be improved without sacrificing clean accuracy.

Third, extending the framework to multimodal settings, including text-image pairs or social media posts, may better reflect real-world misinformation scenarios.

Finally, standardized deployment-aware robustness benchmarks could facilitate fair comparison across detectors and promote more realistic evaluation practices in AI media security research.

{
    \small
    \bibliographystyle{ieeenat_fullname}
    \bibliography{main}
}


\end{document}